# A Comparative Study of Multiple Deep Learning Algorithms for Efficient Localization of Bone Joints in the Upper Limbs of Human Body

Soumalya Bose[1]*, Soham Basu[2]**, Indranil Bera[3], Sambit Mallick[4], Snigdha Paul[4], Saumodip Das[4], Swarnendu Sil[4], Swarnava Ghosh[4], Anindya Sen[4]

[1]FAU Erlangen-Nürnberg, Erlangen, Bavaria, Germany
`soumalya.bose@fau.de`
[2]University of Freiburg, Freiburg im Breisgau, Germany
`soham.basu07@gmail.com`
[3]Jadavpur University, Kolkata, West Bengal, India
[4]Heritage Institute of Technology, Kolkata, West Bengal, India
`anindya.sen@heritageit.edu`

***Abstract.*** This paper addresses the medical imaging problem of joint detection in the upper limbs, viz. elbow, shoulder, wrist and finger joints. Localization of joints from X-Ray and Computerized Tomography (CT) scans is an essential step for the assessment of various bone-related medical conditions like Osteoarthritis, Rheumatoid Arthritis, and can even be used for automated bone fracture detection. Automated joint localization also detects the corresponding bones and can serve as input to deep learning-based models used for the computerized diagnosis of the aforementioned medical disorders. This increases the accuracy of prediction and aids the radiologists with analyzing the scans, which is quite a complex and exhausting task. This paper provides a detailed comparative study between diverse Deep Learning (DL) models – YOLOv3, YOLOv7, EfficientDet and CenterNet in multiple bone joint detections in the upper limbs of the human body. The research analyses the performance of different DL models, mathematically, graphically and visually. These models are trained and tested on a portion of the openly available MURA (musculoskeletal radiographs) dataset. The study found that the best Mean Average Precision (mAP$_{0.5:0.95}$) values of YOLOv3, YOLOv7, EfficientDet and CenterNet are 35.3, 48.3, 46.5 and 45.9 respectively. Besides, it has been found YOLOv7 performed the best for accurately predicting the bounding boxes while YOLOv3 performed the worst in the Visual Analysis test.

---

* denotes equal contribution

# 1 Introduction

## 1.1 Motivation

The manual evaluation of medical images by radiologists is quite a crucial but time-consuming task. Nowadays, many studies aim at the introduction of automation in the health care industry. Although various Deep Learning-based Object Detection models exist at present, it was found that relatively less research answers the question – "Which algorithm to use for the detection of bone joints and when"? This study not only discusses different DL models and their performances but also produces a comprehensive, critical and comparative study to help users determine which models to use depending upon their needs.

## 1.2 Models and Dataset

This research considers 4 different Deep Learning models – YOLOv3 [1], YOLOv7 [2], EfficientDet [3] and CenterNet [4] for bone joint localizations in the upper limbs of the human body, including elbow, finger, shoulder and wrist joints. For widening the scope of study and aiming at diversification, detection models from differently families are considered. These models have been trained on a section of the openly available MURA (musculoskeletal radiographs) dataset [5]. Extensive experimentation has been performed by manually altering some of the hyperparameters of especially the EfficientDet and CenterNet models to best suit our problem statement and obtain optimal results.

We consider Yolov3 as the baseline model for our research. Released in 2018, the YOLOv3 was one of the fastest object detection algorithms of its time and proposed multiple improvements over its predecessors and other contemporary models. The YOLOv3 introduced a backbone network with residual connections, bounding box predictions with objectness scores and most importantly, performed multiscale object detection. CenterNet, EfficientDet and YOLOv7 are some of the latest and state-of-the-art models in object detection with YOLOv7 being released as recently as June 2022.

Detection of joints in medical images is quite challenging due to the limited availability of datasets with good-quality images. We have passed the MURA dataset through multiple image preprocessing steps for filtering out the redundant, noisy and low-quality images. Image Augmentation has also been performed to increase the dataset size.

## 1.3 Novelty of our research

Implementing these models has the scope to deliver efficient, faster and precise medical results that not only benefits the patients but also allows recreational space for the service providers. It is often the case that, from a plethora of DL models researchers find it difficult to choose the one that best fits their interest. Our study addresses the question of which Deep Learning algorithm to use based on user-specific needs.

This research not only serves as a guideline for usage in medical imaging, but also showcases how DL-based models released over the years have evolved in terms of speed, accuracy and novelty. Besides, all these models had been originally trained on datasets with natural images only and their efficacy in other fields is relatively unexplored. In this paper, we assess the effectiveness of these models when employed for a specific medical imaging-based task of joint localization. The performance of these models and their comparison has been discussed on three different fronts – Mathematical, Graphical and Image Visuals.

## 2    Background

Pankhania et al. [6] observed that MRI slices with high thickness as well as overlapping anatomical structures in X-rays create a problem for detection, leading to the AI-related possibilities in musculoskeletal radiology. CNNs were found to be useful due to their convolution layer of "artificial neurons" and better working with imaging data. Radial basis neural networks were found to be better in both training and performance, while Generative adversarial networks (GANs), which generates new data, has not yet been used in musculoskeletal radiology.

Kalmet et al. [7] observed that analyzing medical images can be a time-consuming and difficult process, and can easily be solved by Artificial Intelligence (AI). Automated DL systems have the capacity to detect and classify fractures on radiographs and CT images with higher sensitivity and specificity. It was observed that a system consisting of CNN-based feature extraction modules and an RNN module could be deployed for the detection of osteoporotic vertebral fractures (OVF).

Cheng et al. [8] proposed a Deep Convolutional Neural Network (DCNN) based model to identify hip fractures without feeding region specific images. However, the model failed to explain why certain activations during detection occurred in the wrong site. Additional drawbacks include the difficulty in feeding the whole image into the model for detection as well as narrow region of operation on radiographs only.

Lindsey et al. [9] observed that emergency clinicians, due to their lack of orthopedic expertise are prone to errors while diagnosing fractures. To counter this, A DCNN is employed for fracture detection and localization. This approach generates two types of outputs – a single probability value and a dense conditional probability map. The single probability is a confidence score indicating the presence of fractures in the given radiograph while the other is a heat map, wherein each pixel is a confidence score that indicates whether that region is part of a fracture.

Folle et al. [10] tried to differentiate between rheumatoid arthritis (RA), psoriatic arthritis (PsA), and healthy individuals (HC) based on identification of disease-specific regions in the joints using a neural network. The identification was made on High-resolution peripheral-computed topography (HR-pQCT) data whereas the critical spots were mapped using GradCAM. Convolutional Supervised Auto-Encoder (CSAE) was used for the identification and distinction of the arthritis conditions by performing the detection on hand-joint images. This neural network carried out the detections on the heat-map generated by itself, and worked fine even with the outer contours of the bone joint. This model, however, has few scopes of improvement. Firstly, the HR-pQCT data used in this model is not available in large quantities, and secondly, this model could work only on the rheumatoid arthritis (RA), psoriatic arthritis (PsA), and healthy individuals (HC), and tried to find the best match of any unknown disease to these three conditions only.

Jakaite et al. [11] commented that existing ML models provide limited diagnostic accuracy, thus proposed a new Group Method of Data Handling Strategy of Deep Learning (GMDH). This model works on Zernike-based texture features which improve the diagnostic average by around 11%. This experiment had been conducted using four Machine Learning techniques RF, SVM, ANN, and the proposed GMDH-type network. It was observed that Zernike moments have better precision than Haralick features for all four techniques, while the proposed GMDH method showed better results in comparison to the other three models.

Wang et al. [12] reviewed the various scopes of Deep Learning algorithms in image reconstruction for magnetic resonance imaging (MRI), computed tomography (CT), and positron tomography (PET).

Xue et al. [13] proposed an automated learning-based model that can be used to locate anatomical landmarks on the knee joint using three-dimensional MR image. In training, a Difference of Gaussian (DoG) detector was used for landmark detection, followed by a Scale Invariant Feature Transform (SIFT) descriptor for landmark characterizing and finally a multi-classifier boosting system was deployed for high localization accuracy. This model, based on measuring the distance between manual landmarks and detected landmarks, achieved landmark detection with an accuracy of 70%.

## 3      Materials

### 3.1    Dataset

All the models have been trained and tested using a filtered version of the MURA (musculoskeletal radiographs) dataset consisting of approximately 36,000 images. It is one of the most extensive public radiographic image datasets of bone X-rays collected from the Stanford ML group [5]. The MURA dataset includes 4429, 8395, 9764 and 5106 images of elbow, finger, shoulder and wrist joints respectively, with and without bone fractures.

### 3.2    Preprocessing

The MURA dataset was used for our research, which consists of approximately 36,000 images. However, this dataset also includes images of many other bone joints than the ones we required. As a result, we have executed a series of image preprocessing steps on MURA.

First, image redundancy in the dataset has been removed using the AllDup software, followed by a manual removal of noisy images. Noisy images include images with poor resolution, poor contrast and features which are not clearly distinguishable by the naked eye. This led to a relatively smaller number of shoulder joint images as compared to the other joints. To create a balanced dataset, we have exploited the benefits of image augmentation and increased the number of shoulder joint images through basic image processing techniques like random flipping and random rotation.

These series of image processing steps (see Fig. 1) led to a final, balanced dataset of 4929 images, split into a 70:30 ratio of train and test sets having 3413 and 1516 images respectively. Finally, the images were annotated using the LabelImg software by manually drawing rectangular bounding boxes.

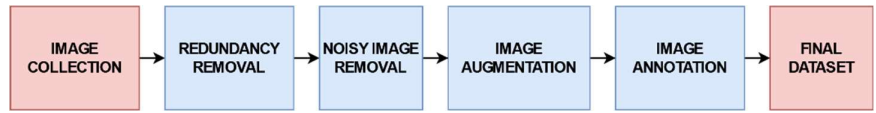

**Fig. 1.** Image Preprocessing Flow

### 3.3 Hardware and Libraries

The codes for model training and inference were written in Python in Jupyter notebooks executed on virtual machines provided by Google Colaboratory. Google Colab's free tier GPU runtimes provide a single Tesla T4 (2560 CUDA cores and 16 GB VRAM) GPU with unstated (approx. 4-6 hours) usage limits.

## 4 Proposed Methods

### 4.1 YOLOv3

**Network Architecture.** YOLOv3 is based on Darknet-53 backbone (Fig. 2). Darknet-53 is a feature extractor composed of 53 convolutional layers with residual connections. Each convolution layer uses Leaky-ReLU activation function and Batch Normalization.

**Fig. 2.** Darknet-53 architecture (reproduced from [1])

**Model Training.** YOLOv3 takes images of size 416×416 pixels as input. The model is trained for a maximum of 8000 iterations with a batch size of 8. The learning rate (LR) is kept at a constant 1e-3 throughout training.

**Loss Functions and Optimizer.** YOLOv3 uses Sum of Squared Error Loss (SSE) for predicting bounding boxes, which is given by:

$$SSE = \frac{1}{m}\sum_{i=1}^{m}|y - \hat{y}|^2 \qquad (1)$$

where $y$ is the predicted value, $\hat{y}$ is the ground-truth value and $m$ is the number of examples. For the class predictions, YOLOv3 employs Multi-class Cross-Entropy Loss ($J_{CE}$), computed by:

$$J_{CE} = -\frac{1}{N}\sum_{i=1}^{N}\sum_{j=1}^{m} y_{ij} \cdot \log\left(p(y_{ij})\right) \qquad (2)$$

where $y_{ij}$ is the actual label, $p(y_{ij})$ is the predicted probability of $y_{ij}$, $m$ is the number of classes and N is the total number of images/examples.

The SGD optimizer is employed, maintaining all the originally used parameters.

### 4.2 YOLOv7

**Network Architecture.** YOLOv7's architecture is based on the extended ELAN or E-ELAN backbone (Fig. 3). This process uses group convolutions to expand the channels as well as the number of computational blocks. The computational blocks are used to calculate feature maps. The blocks are then shuffled and concatenated together, and network depth and width are scaled.

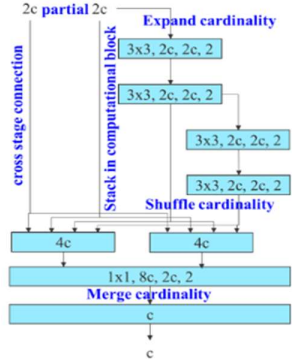

**Fig. 3.** E-ELAN architecture (Image reproduced from [2])

**Model Training**. YOLOv7 takes input images of size 640×640 pixels. We maintain the One Cycle Policy of cycling the learning rate with an initial rate (lr0) 1e-2 and final rate (lrf) 1e-1. A batch size of 8 is used, and the model is trained for 8000 iterations. SGD is employed to optimize the loss functions, maintaining the original parameters.

### 4.3 EfficientDet

**EfficientDet Architecture.** The EfficientDet model comprise the following key features.

1. *EfficientNet Backbone.* EfficientDet uses the EfficientNet family of models as the feature extraction backbone [14]. The paper proposed a new compound model scaling method for uniformly scaling all the dimensions of an image classification model viz. depth, width and resolution. The architecture of the baseline model termed EfficientNet-B0 has been shown in Fig. 4.

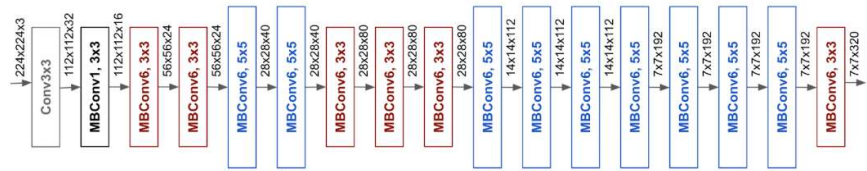

**Fig. 4.** EfficientNet-B0 architecture (Image from Google AI Blog [1]).

2. *Bidirectional Feature Pyramid network (BiFPN).* The BiFPN uses weighted feature fusion and bidirectional cross-scale connections for multi-scale feature fusion.
3. *Compound Scaling.* Building on the scalable nature of EfficientNet, EfficientDet uses a compound scaling method that scales up exponentially the width of the BiFPN, and linearly the depth of the BiFPN, Class/Box prediction networks and input image resolution, using a compound scaling coefficient Φ.

We use EfficientDet-D1 (see Table T1) with Φ = 1. Due to resource constraints, training a further upscaled model was not feasible. The generalized architecture of EfficientDet shown in Fig. 5 has been reproduced from [3].

**Table 1.** Scaled dimensions and layers of EfficientDet-D1 (reproduced from [3])

| Scaling Coefficient | Input Size (pixels) | Backbone | BiFPN | | No. of box/class layers |
|---|---|---|---|---|---|
| | | | No. of channels | No. of layers | |
| D1 (Φ = 1) | 640 | EfficientNet-B1 | 88 | 4 | 3 |

---

[1] https://ai.googleblog.com/2019/05/efficientnet-improving-accuracy-and.html

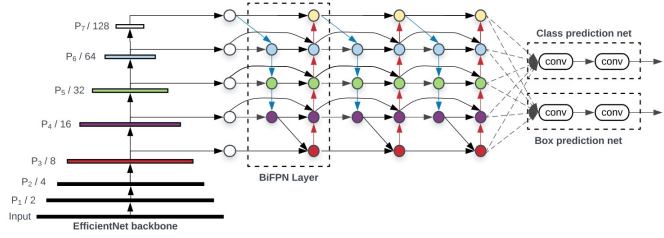

**Fig. 5.** EfficientDet architecture

**Training.** The model is trained for a maximum of 13000 iterations in two separate runs of 8000 and 5000 iterations respectively, with a batch size of 8. In the first run ($i = 1$), the learning rate is linearly increased for the first training epoch (approx. 426 iterations) from 1e-4 to 3e-3 and then annealed down using the cosine decay rule [15], given by:

$$\eta_t = \eta_{min}^i + \frac{1}{2}(\eta_{max}^i - \eta_{min}^i)\left(1 + \cos(\frac{T_{cur} - T_{wp}^i}{T_i - T_{wp}^i}\pi)\right) \quad (3)$$

where $\eta_t$ is resulting learning rate, $\eta_{min}^i$ and $\eta_{max}^i$ are the minimum (annealed final) and maximum (base) values of the learning rate in the $i$th run, $T_{cur}$ is the current iteration and $T_i$ is the maximum number of iterations in the $i$th run. $T_{wp}^i$ is the number of warm-up iterations. In the first run, $T_{wp}^i = 426$.

For the second run ($i = 2$), the base learning rate ($\eta_{max}^i$) is decreased to 3e-4 with $T_i = 17065$. Since $T_{cur} = 8001$, the starting learning rate of this run = 1.647e-4. It is then annealed down using the cosine decay rule. $T_{wp}^i = 1$ for the second run.

**Losses, Loss Function and Optimizer**. The EfficientDet model outputs 3 different losses: Classification Loss and Localization Loss from the class and box prediction networks respectively, and the Regularization loss. Focal Loss [16] is the loss function employed (Eqn. 4) with $\alpha_t = 0.25$ and $\gamma = 1.5$.

$$\text{FL}(p_t) = -\alpha(1 - p_t)^\gamma \log(p_t) \quad (4)$$

where $p_t$ is the estimated probability for the class $y = 1$, $\alpha_t \epsilon$ [0,1] is the weighting factor and $\gamma \geq 1$ is the focusing parameter.

We use the Adam [17] to optimize the total loss.

### 4.4 CenterNet

**CenterNet Architecture.** The architecture of the CenterNet model used has been summarized as follows.

1. *Residual Network Backbone.* We use ResNet-v1-101 [18] as the feature extracting backbone of the network. It is composed of 101 ReLU activated convolution layers with skip connections (Table 2).

**Table 2.** ResNet-v1-101 layers' summary [18]

| Layer name | conv1 | conv2_x | conv3_x | conv4_x | conv5_x | |
|---|---|---|---|---|---|---|
| No. of blocks | 7×7, 64, stride 2 | [1×1, 64; 3×3, 64; 1×1, 256] ×3 | [1×1, 128; 3×3, 128; 1×1, 512] ×4 | [1×1, 256; 3×3, 256; 1×1, 1024] ×23 | [1×1, 64; 3×3, 64; 1×1, 2048] ×3 | Average Pooling, 1000-d Fully Connected, Softmax |
| Output size | 112×112 | 56×56 | 28×28 | 14×14 | 7×7 | 1×1 |
| FLOPs | | | | $7.6 \times 10^9$ | | |

2. *Feature Pyramid Network (FPN).* The output of the ResNet backbone is fed to the FPN that uses bottom-up, top-down and lateral connections in typical CNN feature pyramids, and makes multi-scale object detection predictions independently at the different levels. The conceptual architecture of FPN is illustrated in Fig. 6.

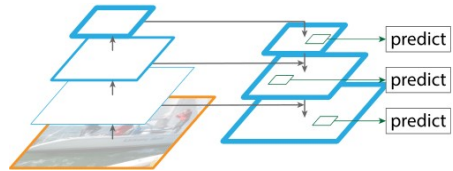

**Fig. 6.** Feature Pyramid Network (reproduced from [19])

3. *Keypoint Prediction.* CenterNet replaces the usual anchor boxes approach with keypoint prediction. It tries to estimate the centers of objects in images and generates heatmaps. Using the center coordinates, bounding boxes are drawn during inference.

**Training.** The input image size is 512×512 pixels. Keeping a batch size of 8, the CenterNet model has been trained for a total of 26000 iterations, in multiple runs lasting a few thousand iterations. The changes in learning rate have been noted in Table 3.

**Table 3.** Variation of Learning Rate while training the CenterNet model.

| Iterations | Warm-up Learning Rate | Warm-up steps | Learning rate |
|---|---|---|---|
| 1 – 16000 | 2.5e-4 | 426 | 1e-3 |
| 16001 – 18000 | - | - | 7e-4 |
| 18001 – 22000 | - | - | 1e-5 |
| 22001 – 26000 | - | - | 1e-4 |

**Losses, Loss Functions and Optimizer**. [4] mentions the use of 3 losses. For the heatmaps generated by the Keypoint Estimation Network, a penalty-reduced Focal Loss [16] is used:

$$L_k = \frac{-1}{N}\sum_{xyc}\begin{cases}(1-\hat{Y}_{xyc})^\alpha \log(\hat{Y}_{xyc}) & if\ Y_{xyc}=1 \\ (1-Y_{xyc})^\beta (\hat{Y}_{xyc})^\alpha \log(1-\hat{Y}_{xyc}) & otherwise\end{cases} \quad (5)$$

where $\hat{Y}_{xyc}$ signifies keypoint prediction: $\hat{Y}=1$ refers to a detected keypoint whereas, $\hat{Y}_{xyc}$ is background. Here, ResNet-101-v1 with FPN is used to predict $\hat{Y}$ from Image $I \in R^{W \times H \times 3}$ of width $W$, height $H$ and $R$ is the output stride. [4] uses output stride $R=4$, which downsamples the output by 4. In Eqn. 5, $N$ is the number of keypoints in the image, and $\alpha=2, \beta=4$ are the hyperparameters of Focal Loss. $Y$ is a Gaussian kernel, $Y_{xyc} = \exp\left(-\frac{(x-\tilde{p}_x)^2+(y-\tilde{p}_y)^2}{2\sigma_p^2}\right)$, where $\sigma_p$ is an object size-adaptive standard deviation, $p \in R^2$ is the ground truth keypoint of class $C$ and $\tilde{p} \in \frac{p}{R}$ is a low-resolution equivalent of $p$. In our case, the keypoint types include $C=4$ classes or the 4 joint types. For class offsets, L1 loss is used:

$$L_{off} = \frac{1}{N}\sum_p \left|\hat{O}_{\tilde{p}} - \left(\frac{p}{R} - \tilde{p}\right)\right| \quad (6)$$

where $\hat{O}$ is the local offset. L1 loss is also used for the regressing the bounding box dimensions:

$$L_{size} = \frac{1}{N}\sum_{k=1}^{N}|\hat{S}_{pk} - S_k| \quad (7)$$

where $\hat{S}_{pk}$ is the predicted size of the bounding box and $S_k$ is the ground-truth bounding box size. Thus, the final training objective loss is given by:

$$L_{det} = L_k + \lambda_{off}L_{off} + \lambda_{size}L_{size} \quad (8)$$

where $\lambda_{off}=1$ and $\lambda_{size}=0.1$ (original values according to [4]).

## 5 Results

### 5.1 YOLOv3

YOLOv3 takes around 2 hours and 8 minutes to train 8000 iterations. The model on evaluation yields the results shown in Table 4 and training loss curve in Fig. 7.

Table 4. Evaluation results of YOLOv3 after 8000 iterations

| mAP$_{0.5}$ | mAP$_{0.75}$ | mAP$_{0.5:0.95}$ |
|---|---|---|
| 85.7 | 18.8 | 35.3 |

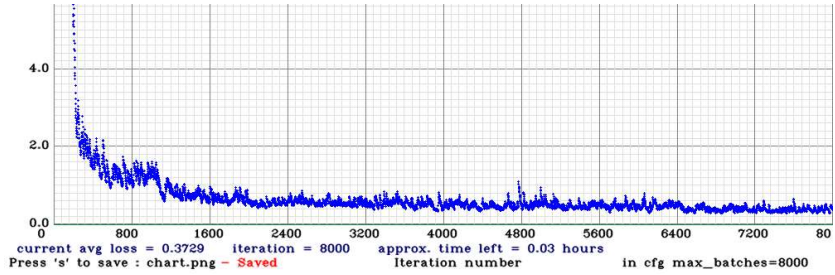

**Fig. 7.** Training Loss Curve of YOLOv3

### 5.2 YOLOv7

The training time of YOLOv7 is almost 1.8 hours for 8000 iterations on 3413 images. Evaluation Results are shown in Table 5:

**Table 5.** Evaluation results of YOLOv7 after 8000 iterations

| $mAP_{0.5}$ | $mAP_{0.5:0.95}$ |
|---|---|
| 91.4 | 48.3 |

The precision vs recall curve of YOLOv7 at 0.5 IoU threshold is shown in Fig. 8, while Fig. 9 plots the train loss against epochs (8000 iterations ≈ 19 epochs).

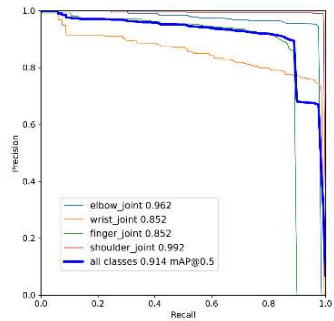 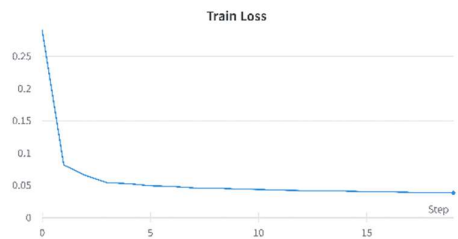

**Fig. 8.** Precision-Recall Curve of YOLOv7

**Fig. 9.** Total Loss Curve of YOLOv7

### 5.3 EfficientDet

EfficientDet-D1 takes almost 3 hours to train 8000 iterations. The evaluation results of the model are tabulated in Table 6 and training loss curve is shown in Fig. 10. Fig. 11a and b show the variation of $mAP_{0.5:0.95}$ after every 1000 training iterations.

**Table 6.** Evaluation results of EfficientDet-D1 with respect to training iterations

| After Training Iterations | mAP$_{0.5}$ | mAP$_{0.75}$ | mAP$_{0.5:0.95}$ |
|---|---|---|---|
| 8000 | 92.0 | 37.0 | 45.1 |
| 13000 | 92.6 | 40.7 | 46.5 |

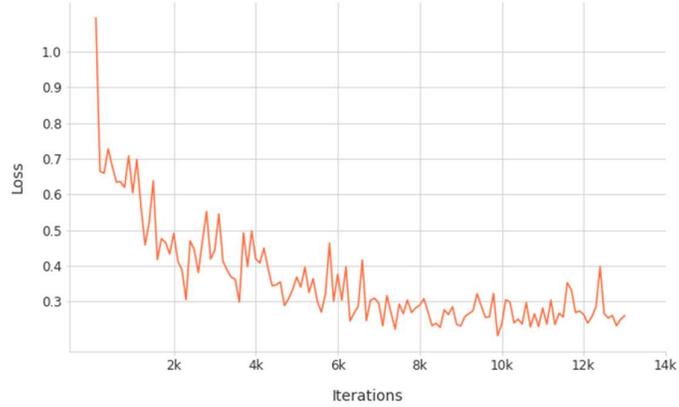

**Fig. 10.** EfficientDet-D1 Training Loss curve

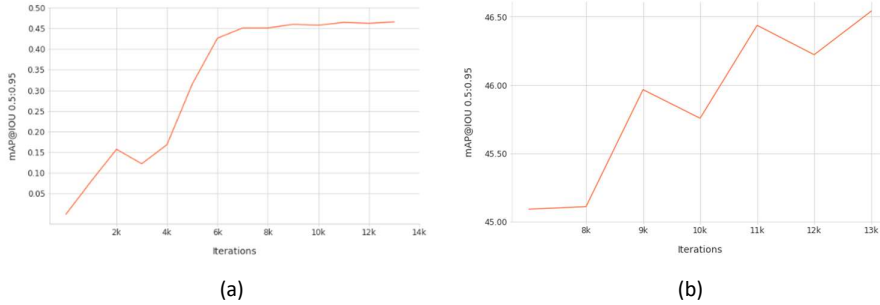

**Fig. 11.** (a). mAP$_{0.5:0.95}$ variations of EfficientDet-D1, (b) enlarged version of (a) to show variations in the 7000 to 13000 iterations range

### 5.4 CenterNet

CenterNet takes around 1.7 hours to train 8000 iterations. Fig. 12 shows the training loss curve of the model. CenterNet with a ResNet-101 backbone and FPN achieves the best mAP$_{0.5:0.95}$ of 45.9 after 19000 training iterations (Fig. 12). The detailed evaluation results of the model, with respect to learning rate have been shown in Table 7. Fig. 13a and b show the variation of mAP$_{0.5:0.95}$ after every 1000 training iterations.

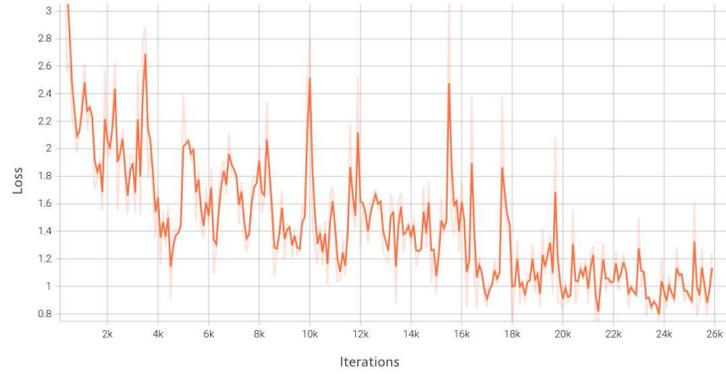

**Fig. 12.** Loss curve of CenterNet with ResNet-101 and FPN

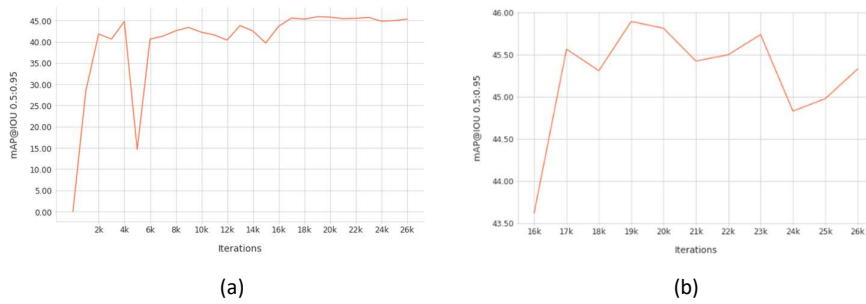

(a)                                                          (b)

**Fig. 13.** (a). mAP$_{0.5:0.95}$ variations of CenterNet, (b) enlarged version of (a) to show variations in the 16000 to 26000 iterations range

**Table 7.** Evaluation results of CenterNet

| After Training Iterations | Learning Rate | mAP$_{0.5}$ | mAP$_{0.75}$ | mAP$_{0.5:0.95}$ |
|---|---|---|---|---|
| 8000  | 1e-3 | 87.5 | 33.5 | 42.6 |
| 13000 | 1e-3 | 90.8 | 34.2 | 43.8 |
| 16000 | 1e-3 | 92.3 | 32.4 | 43.6 |
| 18000 | 7e-4 | 93.7 | 36.0 | 45.6 |
| 22000 | 1e-5 | 93.9 | 35.6 | 45.5 |
| 26000 | 1e-4 | 93.4 | 35.0 | 45.3 |

## 6  Discussion

### 6.1  Mathematical Analysis

Table 8 shows a detailed comparison between the training times for 8000 iterations of the 4 models used in this research. CenterNet and YOLOv7 train the fastest by far, while EfficientDet-D1 takes the longest time to train.

**Table 8.** Detailed comparison of models' performance for 8000 iterations

| Models | $mAP_{0.5}$ | $mAP_{0.75}$ | $mAP_{0.5:0.95}$ | Total training time (hours) | Average training time per iteration (s) |
|---|---|---|---|---|---|
| YOLOv3 | 85.7 | 18.8 | 35.3 | 2.13 | 0.96 |
| YOLOv7 | 91.4 | - | 48.3 | 1.68 | 0.76 |
| EfficientDet-D1 | 92.0 | 37.0 | 45.1 | 3.11 | 1.40 |
| CenterNet | 87.5 | 33.5 | 42.6 | 1.67 | 0.75 |

However, YOLOv7 produces the best results at higher IOU values (as indicated by its highest $mAP_{0.5:0.95}$ of 48.3) after training for only 8000 iterations. Although EfficientDet-D1 trains the slowest, it achieves a decent $mAP_{0.5:0.95}$ value of 45.1 and in fact the highest $mAP_{0.5}$ of 92.0. The authors of [3] train the models for 300 epochs (approx. 128000 iterations) to achieve the best results. However, we have not been able to do the same due to resource constraints. CenterNet and YOLOv3 lag the other two models quite evidently.

In conclusion, YOLOv7 is more suited to users with the low-end hardware resources who would prefer achieving the best $mAP_{0.5:0.95}$ performance by training the least.

**Table 9.** Best mAP comparison between models

| Model | $mAP_{0.5}$ | $mAP_{0.75}$ | $mAP_{0.5:0.95}$ | Training Iterations |
|---|---|---|---|---|
| YOLOv3 | 85.7 | 18.8 | 35.3 | 8000 |
| YOLOv7 | 91.4 | - | 48.3 | 8000 |
| EfficientDet-D1 | 92.6 | 40.7 | 46.5 | 13000 |
| CenterNet | 94.0 | 36.0 | 45.9 | 19000 |

In Table 9 we compare the best obtained mAP values of the 4 models. YOLOv7 still remains the best performing model according to the $mAP_{0.5:0.95}$ metric with a value of 48.3. EfficientDet-D1 on the other hand catches up quite close to YOLOv7's $mAP_{0.5:0.95}$ after training for 5000 iterations more. Training for 13000 iterations results in the increase in $mAP_{0.75}$ by 3.7, i.e., 10%. Though the model is far from convergence, further training might result in EfficientDet-D1 reaching the mAP of YOLOv7.

### 6.2 Graphical Analysis

**Loss curves of YOLOv3 and YOLOv7.** The train loss curves of YOLOv3 (Fig. 7) and YOLOv7 (Fig. 9) show that the models' losses have almost stagnated, indicating that the models could be considered to have nearly converged. Training the models further could lead to overfitting.

**EfficientDet-D1 could be trained for longer.** The loss curve of EfficientDet-D1 shows an overall decrease in loss until the end of 8000 iterations. Although the loss slightly stagnates after that, it can be seen from Fig. 11b that the mAP continues to increase. This signifies that the model could be trained for much longer and better results could be obtained.

**Oscillating nature of CenterNet's loss curve**. It is interesting to note that CenterNet's loss curve oscillates a fair amount during training (Fig. 12). Although some oscillations are expected due to the small batch size, the magnitude of fluctuations is too high to be ignored. Following [4], we train the model for 16000 iterations during which we notice not only an oscillating loss curve but also oscillations in the mAP values after evaluation (Fig. 13a). To overcome this, we decrease the learning rate to 7e-4 which seems to dampen the oscillations a little and leads to a steep rise in the mAP values. The LR is further decreased to 1e-5 and the best $mAP_{50:95}$ value of 45.9 is obtained at the end of the 19000th iteration. The LR is now increased to 1e-4 to test the presence of saddle points, but the mAP further drops down after the 23000th iteration. This indicates that an LR of around 1e-5 may lead to the best evaluation results on further training.

### 6.3 Visual Analysis

Fig. 13, 14, 15, 16: [2](a), [3](b), [4](c), [5](d) and [6](e) show the performance of the YOLOv3, YOLOv7, EfficientDet-D1 and CenterNet models respectively, on 5 images obtained from the internet. These images contain multiple joints and/or multiple occurrences of the same joint. They contain watermarks, noise, fractures and are of low quality – factors which help determine the robustness of the employed models. The inference results clearly outline the superiority of the YOLOv7 model which predicts most of the joints in the images correctly and with high confidence scores, followed by EfficientDet-D1 and CenterNet, while YOLOv3 is barely able to detect a few joints.

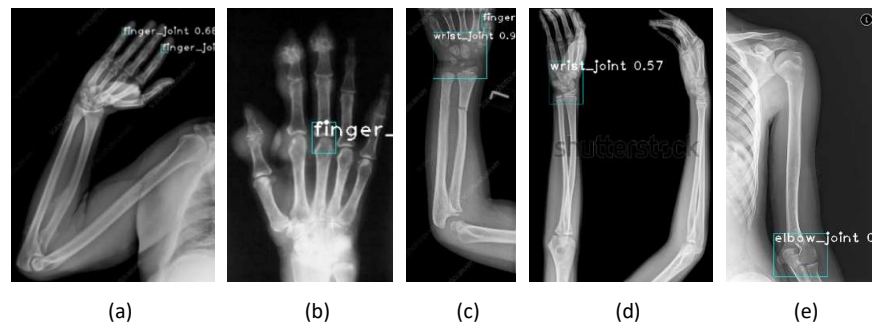

(a)　　　　(b)　　　　(c)　　　　(d)　　　　(e)

**Fig. 14.** Inference results of YOLOv3.

---

[2] https://www.sciencephoto.com/media/769442/view/healthy-arm-x-ray
[3] https://img.medscapestatic.com/pi/meds/ckb/94/19894tn.jpg
[4] https://media.sciencephoto.com/image/c0177264/800wm/C0177264-Broken_arm,_X-ray.jpg
[5] https://www.shutterstock.com/image-photo/xray-both-human-arms-hands-600w-227195635.jpg
[6] https://upload.wikimedia.org/wikipedia/commons/3/3e/Medical_X-Ray_imaging_MXK06_nevit.jpg

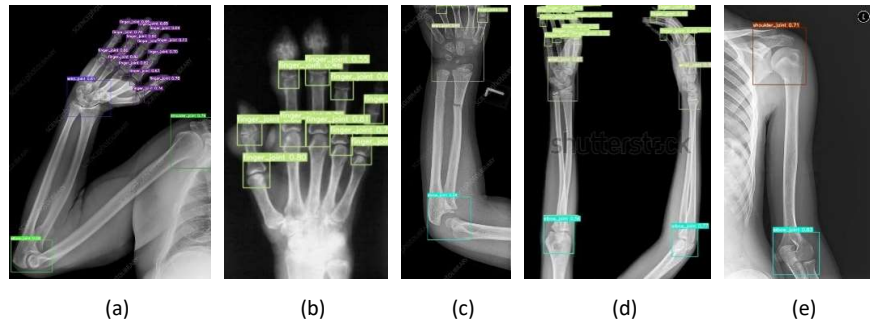

**Fig. 15.** Inference results of YOLOv7.

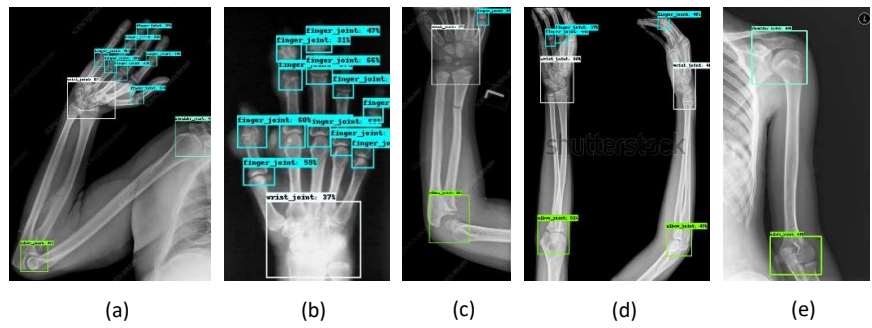

**Fig. 16.** Inference results of EfficientDet-D1.

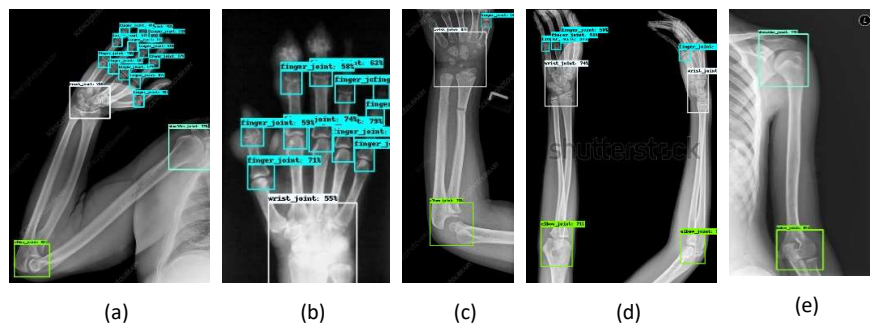

**Fig. 17.** Inference results of CenterNet.

# 7 Conclusion

In this paper we train and evaluate deep learning-based object detection models on Stanford's MURA dataset for detecting bone joints in images. Our experimental results show that the selected models achieve noteworthy performance. These results suggest that even with a dearth of data, modern DL-based detection models used for natural images are also quite effective when trained and evaluated using medical datasets. However, we feel that there are certain shortcomings that also need to be addressed.

Due to resource constraints, we have only been able to train the EfficientDet-D1 model and have not been able to consider the best performing and highest scaled EfficientDet-D7 and D7x. These two models could be expected to significantly outperform the EfficientDet-D1 model and perhaps even YOLOv7.

Similarly, we have not been able to consider YOLOv7-E6E which is the fastest and best performing YOLOv7 model on the Microsoft COCO[7] test-dev dataset.

The two YOLO versions used in this paper have significant differences. YOLOv3 uses the old Darknet-53 backbone while YOLOv7 employs the novel Extended ELAN backbone and features a scalable architecture like EfficientDet. Although the YOLOv7 performs the best in this study, it must be noted that the model is relatively new and further improvements are always around the corner.

The CenterNet model not confirming to the training strategies suggested by Zhou et al. [4] indicates that a better training strategy needs to be adapted. Besides, the same relative performance of the model on the COCO test-dev dataset cannot be assumed in this problem of medical imaging. Further training and experimentation with varying learning rates could result in much higher mAP values.

The backbone of the CenterNet model used in this research – ResNet-101-v1 could also be replaced by a better performing classification architecture like ResNeXt-101 [20]. Similarly, the BiFPN proposed by [3] could also replace the FPN used here, as it is an improvement over the latter.

The scope of object detection models is ever-expanding. Most of the effort in deep learning research is trial and error, and reuse of neural network blocks (e.g., Residual blocks, bottlenecks and FPN) that could possibly improve performance. This, applied to medical imaging, could also help expand the applicability and accuracy of radiological techniques for the diagnosis of musculoskeletal diseases as well as injuries.

Availability of high-quality data in medical imaging is often scarce. We feel that the size of the current dataset could be further increased for better training of and generalization by the models.

Two-stage detection models such as Faster-R-CNN, R-FCN and Cascade R-CNN have not been included in this study as the authors faced hardware constraints while training. Training on a single Tesla T4 provided by Google Colab has been taking a significant amount of time. In future work, with higher-end GPU and without usage limits, we could test the performance of the aforementioned two-stage detectors.

---

[7] https://cocodataset.org/

This research was carried without any large-scale variations of the built-in hyperparameters like learning rates of the YOLOv3 and YOLOv7 models. Hence, the effects of such changes are unexplored.

## Authors' Contributions

Principal Investigator and Faculty member: A. Sen; conception of presented idea: S. Ghosh; research supervision, primary manuscript writing: S. Bose and S. Basu; manuscript formatting, major proofreading, EfficientDet and CenterNet implementations and experimentations: S. Basu; implementation of YOLOv3 and experiment supervision: S. Bose and I. Bera; YOLOv7 implementation, YOLOv3 and YOLOv7 experimentations and simulations: S. Mallick and S. Paul; background literature study and labelling of images: S. Das and S. Sil; dataset collection, image preprocessing and final dataset preparation: S. Ghosh and I. Bera; interpretation and analysis of research results: A. Sen, S. Basu, S. Bose, S. Mallick and S. Paul. All authors have read and approved the final manuscript.

## Conflict of Interest

The authors have no conflicts of interest, financial or otherwise.

## Information about Funding

The authors did not receive any financial support in the form of grants.

## Resource Availability

Colab Notebooks, figures and weight files used in this research and results are available at: https://github.com/Sohambasu07/BoneJointsLocalization